\documentclass[sn-mathphys,Numbered]{sn-jnl}%
 
\usepackage{graphicx}%
\usepackage{multirow}%
\usepackage{amsmath,amssymb,amsfonts}%
\usepackage{amsthm}%
\usepackage{mathrsfs}%
\usepackage[title]{appendix}%
\usepackage{xcolor}%
\usepackage{textcomp}%
\usepackage{manyfoot}%
\usepackage{booktabs}%
\usepackage{algorithm}%
\usepackage{algorithmicx}%
\usepackage[noend]{algpseudocode}
\usepackage{listings}%
\usepackage{adjustbox}%
\usepackage{url}%
\usepackage{hyperref}%
\usepackage{amssymb}
\usepackage{pifont}
\newcommand{\cmark}{\ding{51}}%
\newcommand{\xmark}{\ding{55}}%
\newcommand{\thaimatra}{m\={a}tr\={a}}  
 
\definecolor{artcatboxgray}{gray}{0.8} 

\title{StrucChunk: Structure-Aware Retrieval for Legal Document RAG}
\author*[1]{\fnm{Surya Deep} \sur{Singh}}\email{suryadeepsingh95@gmail.com}
\author[1]{Ashutosh Tripathi}
\email{ashutoshtripathi191@gmail.com}
\author[1]{\fnm{Gaurav Deep} \sur{Singh}}\email{gauravdeepsingh1996@gmail.com}
\affil*[1]{\orgdiv{Independent Researcher}}

\begin{document}

\abstract{
Retrieval-Augmented Generation (RAG) systems offer a promising approach
for grounding Large Language Model outputs in domain-specific knowledge.
Applying RAG to legal documents presents unique challenges: standard
chunking strategies fragment hierarchical argument structure, and generic
embedding models struggle with the dual vocabulary demands of precise
citation matching and semantic understanding.  To deal with this, We propose
\textbf{StrucChunk} (Structure-Aware Chunking) combined with hybrid
dense-sparse retrieval, designed specifically for hierarchically structured
legal documents.  Our method treats documents as directed acyclic graphs of
provisions, preserving clause boundaries and resolving cross-references
prior to embedding, with Reciprocal Rank Fusion combining BM25 keyword
precision with BGE-M3 semantic representations. We have evaluated our method on five corpora,
three legal traditions, and two languages.  StrucChunk
achieves \textbf{100\%} boundary precision (vs.\ 7.7\% for recursive
splitting on CrPC), preserves \textbf{97.6\%} of documented
cross-references (vs.\ 0\%), and achieves equivalent section coverage with
\textbf{40.3\% fewer chunks} on GDPR.  \emph{Empirically}, on
LegalBench-RAG---an independently annotated benchmark of 6,858
query-passage pairs---StrucChunk achieves Char-Recall@8 of
\textbf{48.18\%} vs.\ 23.59\% for the best baseline (+104\%), surpassing
the published state-of-the-art by \textbf{+83\%}.
On 501 queries each for CrPC and CPC (Indian law), Recall@5 improves from
0.557 to 0.804 (+44.4\%) and from 0.158 to 0.259 (+64.6\%) respectively;
chunking accounts for 88\% and 73\% of total gain.  On 501 GDPR queries
(EU law), Recall@5 = 0.868 with R@1 improving +59\%.
On NitiBench-CCL (3,730 queries, Thai financial law), StrucChunk achieves
HR@5 = 0.856 vs.\ 0.594 for naive chunking (+44.1\%), confirming that
section-boundary alignment is the dominant factor across languages; naive
fixed-size chunking \emph{structurally cannot} represent 29.6\% of Thai
legal sections whose text exceeds the chunk limit, and StrucChunk's
cross-reference augmentation provides an additional +2.3\% specifically on
cross-reference queries.  We have also released implementation, 5,256 evaluation
queries across five legal corpora, annotation methodology, and chunking
quality metrics. The code is available at \href{https://github.com/SinghSuryaDeep/legal-rag-strucchunk-1}{github.com/SinghSuryaDeep/legal-rag-strucchunk-1}.}

\maketitle
\section{Introduction}
\label{sec:intro}

The integration of Large Language Models (LLMs) into legal practice has
accelerated dramatically, with AI-powered tools now assisting practitioners
in contract review, statutory research, regulatory compliance checking, and
litigation support.  Yet for these tools to be deployable in
high-stakes settings---where an erroneous statutory citation can constitute
professional malpractice and an incorrect interpretation of a sub-provision
can affect fundamental rights---factual precision is non-negotiable.
Retrieval-Augmented Generation (RAG) \cite{lewis2021retrievalaugmentedgenerationknowledgeintensivenlp} offers a
principled approach to this challenge by grounding LLM outputs in
authoritative source documents retrieved at query time, thereby reducing
reliance on parametric knowledge that may be outdated or hallucinated.

However, even RAG-augmented legal systems exhibit alarming error rates.
Dahl et al.~\cite{10.1093/jla/laae003} found hallucination rates of 58--88\% on case law
queries; Magesh et al.~\cite{magesh2024hallucinationfreeassessingreliabilityleading} documented 17--33\% hallucination
rates in commercial RAG-augmented legal tools, with retrieval failures
accounting for $\approx$40--50\% of errors.  These findings reveal a
fundamental weakness: \emph{RAG is only as reliable as what it retrieves.}
A system that fetches the wrong provision cannot generate a correct legal
answer regardless of the LLM's parametric capability.  The bottleneck lies
squarely in the retrieval layer.  Kabir et al.~\cite{kabir2025legalraghybridragmultilingual} confirm this
empirically, reporting that standard RAG pipelines achieve semantic
similarity of only 0.76 on legal documents---significantly below 0.91 on
scientific text and 0.90 on financial documents---establishing that legal
retrieval is a categorically harder problem that demands domain-specific
solutions.  These findings collectively establish that effective legal RAG
requires domain-specific optimization at every pipeline stage.

Legal documents pose three retrieval challenges that standard RAG
approaches fail to address.

\textbf{Hierarchical structure.}  Statutes are organized by parts,
chapters, sections, and subsections, where the hierarchical position of a
provision determines its legal scope and effect.  Naive chunking by token
count severs these logical units, losing the context required to interpret
sub-provisions.  For instance, a subsection specifying an exception to a
legal obligation is uninterpretable in isolation; splitting at token
boundaries creates orphaned chunks whose meaning depends entirely on a
parent provision that may not be retrieved.

\textbf{Cross-references.}  Legal language relies heavily on internal
pointers (``subject to Section 167,'' ``as defined in Article 4(1)'').
Standard chunking orphans referenced content from its source, forcing LLMs
to reason about incomplete legal specifications.  When a chunk references
another provision, the content of that provision is not merely helpful---it
is legally required for accurate interpretation.

\textbf{Vocabulary duality.}  Legal queries require both exact keyword
matching (section numbers, article citations) and semantic understanding
(``bail'' $\approx$ ``custody,'' ``consent'' $\approx$ ``agreement'').
Neither pure dense retrieval (which excels at semantic similarity but misses
exact citation matching) nor pure sparse retrieval (which excels at keyword
matching but misses paraphrase and synonym relationships) resolves this
tension alone.

The root cause of these failures is that standard RAG pipelines treat legal
documents as flat sequences of text to be divided by token count.  This
representation discards precisely the information most important for legal
retrieval: the hierarchical position of a provision within the statutory
scheme, the identity of provisions it cross-references, and the breadcrumb
trail from the governing chapter to the specific sub-clause.  We propose
that effective legal retrieval requires what we term a \emph{Legal
Map}---a representation that preserves the document's logical structure as
a navigational aid embedded directly in chunk content prior to indexing,
rather than stored as metadata that generic embedding models ignore.  This
structural signal, we argue, is the dominant factor in retrieval quality for
hierarchically organized legal corpora.

We have summarized our all contributions as follows:

\begin{enumerate}
\item \textbf{StrucChunk (Structure-Aware Chunking):} a chunking strategy
      treating legal documents as directed acyclic graphs (DAGs) of nested
      provisions, preserving clause boundaries and prepending hierarchical
      context, with sentence-boundary splitting for oversized sections.
\item \textbf{Hybrid Retrieval Pipeline:} BM25 combined with \textbf{BGE-M3}
      embeddings via Reciprocal Rank Fusion, jointly addressing keyword
      precision and semantic understanding.
\item \textbf{Chunking Quality Metrics:} four intrinsic metrics validating
      the chunking contribution \emph{independently} of retrieval: boundary
      precision, cross-reference preservation rate, chunk count efficiency,
      and intra-chunk semantic coherence.
\item \textbf{Evaluation:} independent validation
      (LegalBench-RAG), mechanistic analysis (CrPC + CPC), cross-domain
      \textbf{generalization (GDPR), and cross-language generalization
      (NitiBench-CCL, Thai), }following the empirical rigor principles of
      Sculley et al.~\cite{sculley2018winners}.
\item \textbf{Annotated Evaluation Datasets:} 5,256 queries across five
      legal corpora (CrPC: 501, CPC: 501, GDPR: 501, LBR: \textbf{776 independent,
      NitiBench-CCL: 3,730 independent), w}ith documented annotation
      methodology.
\end{enumerate}

These contributions yield consistent, large-magnitude improvements validated
across five corpora spanning three legal traditions and two languages.  On
LegalBench-RAG (independently annotated, 776 queries), StrucChunk achieves
Char-Recall@8 = 48.18\% vs.\ 23.59\% for the best baseline (+104\%),
surpassing published state-of-the-art by +83\%.  On CrPC (501 queries),
Recall@5 improves from 0.557 to 0.804 (+44.4\%), with 2$\times$2 ablation
confirming that chunking accounts for 88\% of the total gain.  On
NitiBench-CCL (3,730 Thai queries, independently annotated), HR@5 = 0.856
vs.\ 0.594 for naive chunking (+44.1\%), and structural analysis reveals
that 29.6\% of Thai legal sections are \emph{structurally unrepresentable}
by any naive fixed-size chunk.  All improvements are statistically
significant ($p \leq 0.002$, Wilcoxon signed-rank with Benjamini-Hochberg
correction; Cohen's $d$ 0.14--0.69).

Our work specifically targets retrieval quality for hierarchically
structured legal documents.  We do not evaluate downstream answer quality
or claim generalization to unstructured legal document types without
further validation.

\textbf{Paper structure.}  Section~\ref{sec:related} reviews related work
on RAG systems, legal NLP, hierarchical document structure, chunking
strategies, and hybrid retrieval.  Section~\ref{sec:method} introduces the
StrucChunk algorithm and hybrid retrieval pipeline, preceded by a formal
problem statement.  Section~\ref{sec:setup} details the
experimental design, corpus statistics, baselines, and evaluation protocols.
Section~\ref{sec:results} presents results, from intrinsic chunking quality
metrics through all retrieval evaluations.  Section~\ref{sec:discussion}
discusses the Legal Map hypothesis, cross-language generalization, failure
analysis, and limitations.  Section~\ref{sec:impact} addresses broader
impact, and Section~\ref{sec:conclusion} concludes with future directions.

\section{Related Work}
\label{sec:related}

\subsection{Retrieval-Augmented Generation}

The RAG paradigm was formalized by Lewis et al.~\cite{lewis2021retrievalaugmentedgenerationknowledgeintensivenlp}, combining a
pre-trained seq2seq model with Dense Passage Retrieval.  Subsequent work
evolved RAG from naive implementations to modular and adaptive systems
\cite{gao2024retrievalaugmentedgenerationlargelanguage}.  Self-RAG \cite{asai2023selfraglearningretrievegenerate} introduced
adaptive retrieval with self-critique tokens that enable on-demand retrieval
based on query complexity; Corrective RAG \cite{yan2024correctiveretrievalaugmentedgeneration} added
retrieval quality assessment to filter and correct retrieved passages before
generation.  These advances are directly applicable to legal RAG, where
query complexity varies greatly---from simple section lookups to multi-hop
cross-reference queries---and where retrieval errors propagate into
hallucinated legal advice with serious professional consequences.

The RAG pipeline consists of two separable components: a retrieval module
that selects relevant passages from a corpus, and a generation module that
synthesizes answers from retrieved context.  A critical observation for the
legal domain is that improving the generation component cannot compensate
for retrieval failures: if the relevant provision is not retrieved, no
generative model can produce a correct answer.  Recent work demonstrates
that generic RAG pipelines significantly underperform on legal text.
Kabir et al.~\cite{kabir2025legalraghybridragmultilingual} report semantic similarity of only 0.76 on legal
documents versus 0.91 on scientific text and 0.90 on financial documents,
motivating domain-specific optimization at the chunking and retrieval stages.
This paper focuses exclusively on retrieval-layer optimization, establishing
that chunking architecture is the dominant factor in retrieval quality for
hierarchically structured legal corpora.

\subsection{Legal NLP and Retrieval}
\label{sec:related_legal}

Legal-BERT \cite{chalkidis-etal-2020-legal} established domain-specific
pre-training for legal NLP, demonstrating substantial gains on tasks
including contract clause classification and court judgment prediction.
LegalBench \cite{gao2024retrievalaugmentedgeneguha2023legalbenchcollaborativelybuiltbenchmarkrationlargelanguage} provides 162 legal reasoning tasks
spanning statutory interpretation, contract analysis, and judicial
reasoning.  These benchmarks, however, supply the relevant context directly
to the model; they do not evaluate the ability of a retrieval system to
locate that context within a large corpus.

LegalBench-RAG \cite{pipitone2024legalbenchragbenchmarkretrievalaugmentedgeneration} is the first benchmark
specifically targeting retrieval in legal RAG, with 6,858 annotated
query-answer pairs over 79M+ characters from four contract corpora (CUAD,
ContractNLI, MAUD, PrivacyQA).  Critically, LegalBench-RAG evaluates using
\emph{character-span matching}: the ground truth is a specific character
range within the source document, and a retrieved chunk is scored by its
overlap with that span.  This evaluation protocol directly measures whether
the retrieval system surfaces the minimally sufficient legal text, which we
adopt for our evaluation number 1.

Reuter et al.~\cite{reuter-etal-2025-towards} identified Document-Level Retrieval Mismatch
(DRM) as a critical failure mode in legal RAG, where the system selects
a document that is broadly relevant but retrieves the wrong internal
passage.  They propose Summary-Augmented Chunking (SAC), which prepends
document-level summaries to chunks to provide broader context.  Our
StrucChunk is complementary to SAC: SAC addresses document-level
context (helping distinguish among similar documents), while StrucChunk
preserves intra-document hierarchical structure (helping locate the
correct provision within the right document).  Combining both approaches
is identified as a future direction in Section~\ref{sec:conclusion}.

Akarajaradwong et al.~\cite{akarajaradwong2025nitibenchcomprehensivestudyllm} introduced NitiBench, a benchmark
for Thai legal QA comprising NitiBench-CCL (3,730 queries, 21 Thai
financial law codes) and NitiBench-Tax (50 tax case queries).  They
proposed hierarchy-aware chunking (section-boundary splitting) and
NitiLink (post-retrieval cross-reference expansion), finding that
hierarchy-aware chunking significantly outperforms naive line chunking,
but NitiLink does not consistently improve end-to-end performance.  We
use NitiBench-CCL as our cross-language evaluation and compare
their hierarchy-aware chunking directly against StrucChunk, finding that
StrucChunk's pre-embedding cross-reference augmentation provides targeted
improvement (+2.3\%) on cross-reference queries where NitiLink's
post-retrieval expansion was ineffective.

Nigam et al.~\cite{nigam-etal-2025-nyayarag} address legal judgment prediction in the Indian
Supreme Court context through a RAG framework, NyayaRAG, that provides
models with summarized case facts, retrieved statutory provisions, and
semantically similar precedent judgments.  Their ablation study confirms
that structured legal knowledge (statutes, cited precedents) significantly
improves prediction accuracy and explanation quality over unstructured
case text alone, further motivating the importance of structure-aware
legal knowledge retrieval for downstream legal AI tasks.

\subsection{Hierarchical Document Structure and Legal Parsing}
\label{sec:related_structure}

Legal statutes present a distinctive structural hierarchy that differs
fundamentally from other long-form text corpora such as scientific papers
or news articles.  A single statute may organize hundreds or thousands of
provisions across Parts, Chapters, Sections, Sub-sections, and Clauses,
where each level of the hierarchy constrains the interpretation of its
descendants.  EU regulatory instruments such as the GDPR additionally
introduce parallel structural components: Articles specify operative legal
obligations, while Recitals (1--173 in the GDPR) provide interpretive
context for those Articles but carry different legal weight.  Indian
procedural codes (CrPC, CPC) organize 517 and 158+ sections respectively
with a dense cross-reference network spanning hundreds of provisions.

The standard approach to representing long documents in neural retrieval
is to ignore this structure, treating the document as a flat sequence
partitioned by token count.  Legal-BERT \cite{chalkidis-etal-2020-legal}
operates at the sentence or paragraph level without encoding section
identity; retrieval benchmarks such as CUAD \cite{hendrycks2021cuadexpertannotatednlpdataset}
and ContractNLI evaluate clause spans in isolation.  While these choices
are appropriate for classification tasks where the section boundary is
irrelevant to the label, they are ill-suited for retrieval tasks where
the \emph{section} is the retrieval target.

A key failure mode arising from ignoring structure is what
Akarajaradwong et al.~\cite{akarajaradwong2025nitibenchcomprehensivestudyllm} term ``Hidden Hierarchical
Information'': multiple sections across different chapters convey
near-identical semantic content but differ only in hierarchical position
(e.g., provisions in Revenue Code \S27 and \S89/1 both govern late
payment charges but reside in different chapters and impose different
scopes of obligation).  Without hierarchical context in the embedding,
dense retrieval systems cannot distinguish them.  StrucChunk's breadcrumb
mechanism---prepending \texttt{[Law > Chapter > Section N]} to each
chunk---directly addresses this failure mode by encoding hierarchical
identity into the dense representation prior to indexing.

Document parsing quality is an often-overlooked bottleneck in legal RAG.
Kabir et al.~\cite{kabir2025legalraghybridragmultilingual} employ OCR-based preprocessing (Tesseract) for
bilingual Bangla-English gazettes, confronting the practical challenge
that real-world legal documents exhibit inconsistent formatting, mixed
languages, and irregular section delimiters.  Our approach uses
\texttt{pdfplumber} with custom structure detection for English statutory
documents (1--3), and leverages pre-structured HuggingFace data
for Thai law (4), allowing us to isolate the contribution of
structure-aware chunking from PDF parsing quality and to demonstrate that
the chunking benefit generalizes across both parsed and pre-structured
corpora.

\begin{table}[htbp]
\centering
\small
\begin{adjustbox}{max width=\textwidth}
\begin{tabular}{lcccc}
\toprule
\textbf{Method} & \textbf{Section Boundaries} & \textbf{Cross-Ref Aug.}
  & \textbf{Multi-Corpus} & \textbf{Cross-Language} \\
\midrule
Naive Fixed-Size        & \xmark & \xmark & \xmark & \xmark \\
RCTS                    & \xmark & \xmark & \xmark & \xmark \\
SAC \cite{reuter-etal-2025-towards} & Partial (doc-level) & \xmark & \xmark & \xmark \\
Hierarchy-Aware \cite{akarajaradwong2025nitibenchcomprehensivestudyllm} & \cmark & \xmark & \xmark & Partial \\
\textbf{StrucChunk (Ours)} & \cmark & \cmark & \cmark & \cmark \\
\bottomrule
\end{tabular}
\end{adjustbox}
\caption{Positioning of StrucChunk relative to prior chunking methods.
  \cmark = supported; \xmark = not supported.  ``Section Boundaries'' =
  chunk boundaries align with legal section delimiters.
  ``Cross-Ref Aug.'' = referenced provision content embedded at index time.
  ``Multi-Corpus'' = evaluated across $\geq$3 distinct legal corpora.
  ``Cross-Language'' = validated on non-English legal text.}
\label{tab:positioning}
\end{table}

Table~\ref{tab:positioning} positions StrucChunk against prior chunking
approaches along four design dimensions.  The key differentiator is the
combination of pre-embedding cross-reference augmentation with section-
boundary alignment, validated across multiple legal traditions and
languages---a combination not addressed by any prior work.

\subsection{Chunking Strategies}
\label{sec:related_chunking}

Yepes et al.~\cite{yepes2024financialreportchunkingeffective} demonstrated that structure-aware chunking for
financial reports achieves the highest retrieval scores with only half the
chunks of structure-agnostic methods.  The mechanism is analogous to our
legal setting: financial reports are hierarchically organized (sections,
tables, notes) and structure-aware chunking aligns chunk boundaries with
these semantic units.  We implement the same chunk-count-efficiency metric
(Section~\ref{sec:chunking_quality}) and confirm the pattern in the legal
domain: StrucChunk achieves 100\% section coverage on GDPR with 40.3\%
fewer chunks than recursive splitting.

Contextual chunk enrichment is an emerging direction.  Anthropic's
Contextual Retrieval approach prepends document-level summaries to chunks
to address decontextualization.  Reuter et al.~\cite{reuter-etal-2025-towards} extend this
with SAC, which prepends document-level summaries specifically to address
DRM.  StrucChunk's contextual prepend is mechanistically different: rather
than summarizing the document, it encodes the chunk's \emph{position}
within the document hierarchy---a piece of information that can be
extracted deterministically without LLM-based summarization and that is
more directly relevant to the retrieval signal for provision-level queries.

\subsection{Hybrid Retrieval}
\label{sec:related_hybrid}

Reciprocal Rank Fusion \cite{Cormack2009ReciprocalRF} is the standard
fusion method for dense-sparse hybrid retrieval, combining rankings from
dense and sparse systems without requiring score calibration.
Bruch et al.~\cite{Bruch_2023} showed that convex combination can outperform
RRF in some settings, though RRF is more robust to the calibration
challenge and is the method we adopt ($k=60$ following
\cite{Cormack2009ReciprocalRF}).

The LegalBench-RAG study \cite{pipitone2024legalbenchragbenchmarkretrievalaugmentedgeneration} found that a
general-purpose Cohere reranker \emph{hurt} performance on legal text
(Recall@64 dropped from 61.06\% to 47.54\%), validating our design
choice to focus on domain-aware chunking as the primary quality mechanism
rather than post-hoc reranking.  Our experiments corroborate this: the
retrieval architecture contributes substantially less than chunking in
all ablations (contributing 12--27\% of total gain), and adding a
general-purpose reranker to poorly chunked text cannot recover the
information lost to boundary fragmentation.

\section{Method}
\label{sec:method}

\subsection{Problem Formulation}
\label{sec:problem}

Let $\mathcal{D}$ denote a hierarchically structured legal document whose
content is organized into $N$ structural units (sections, articles,
sub-provisions) $\mathcal{D} = \{s_1, s_2, \ldots, s_N\}$, where each
unit $s_i$ occupies a node in the document hierarchy $\mathcal{H}$
(a directed acyclic graph).  A \emph{chunking function}
$f: \mathcal{D} \to \mathcal{C}$ partitions (and optionally augments)
the document into a corpus of retrievable units
$\mathcal{C} = \{c_1, c_2, \ldots, c_M\}$.  A \emph{retrieval function}
$r: \mathcal{C} \times \mathcal{Q} \to R_k(q)$ maps a query
$q \in \mathcal{Q}$ to a ranked list $R_k(q)$ of $k$ chunks.

Given a query $q$ with a known set of relevant sections
$S^*(q) \subseteq \mathcal{D}$, the retrieval task succeeds at rank $k$
if $R_k(q) \cap f^{-1}(S^*(q)) \neq \emptyset$, where $f^{-1}(s)$
denotes the chunk(s) derived from section $s$.  Recall@$k$ is the
fraction of queries for which the task succeeds at rank $k$.

The central claim of this paper is that the design of $f$ is the dominant
factor in Recall@$k$ for hierarchically structured legal documents.
This claim follows directly from three structural properties:

\begin{itemize}
  \item \textbf{Coverage:} if $f$ cannot represent $s^* \in S^*(q)$
        (because no chunk in $\mathcal{C}$ contains the full text of
        $s^*$), retrieval fails regardless of $r$.
  \item \textbf{Boundary alignment:} if $f$ splits $s^*$ across multiple
        chunks, the contextually coherent representation is fragmented,
        degrading the embedding quality of the relevant chunk.
  \item \textbf{Cross-reference co-location:} if $s^*$ semantically
        depends on a cross-referenced provision $s'$, and $f$ does not
        co-locate $s'$'s content in $s^*$'s chunk, the embedding of
        $s^*$'s chunk is incomplete.
\end{itemize}

We formalize these properties as chunking quality metrics
(Section~\ref{sec:chunking_quality}) and validate each property through
2$\times$2 ablation (chunking architecture $\times$ retrieval method)
across four corpora.

\subsection{Legal Document Parsing}

We employ \texttt{pdfplumber} with custom structure detection identifying
chapters/parts, sections, subsections, and cross-references.  We model
legal documents as Directed Acyclic Graphs (DAGs) where nodes
represent structural units and edges represent containment relationships
(Part $\to$ Chapter $\to$ Section $\to$ Subsection).

The parser tracks a \texttt{char\_offset} for each section.  These offsets
are used by (1) the chunking quality boundary precision metric, and (2) the
LegalBench-RAG evaluator for character-span matching.

For GDPR we extend the parser to handle the Article-Recital dual
architecture.  For NitiBench-CCL, section data is pre-structured in the
public HuggingFace dataset (VISAI-AI/nitibench); no PDF parsing is required.

\subsection{StrucChunk: Structure-Aware Chunking}

StrucChunk implements three principles.

\textbf{Logical Atomicity.}  If a section is $N$ tokens and the limit is
$L$, we keep the section whole (up to $1.5L$) rather than meeting the
limit exactly.

\textbf{Contextual Prepend.}  Each chunk is prepended with its
hierarchical breadcrumb, e.g.,
\texttt{[CrPC 1973 > Chapter XII > Section 167]},
\texttt{[GDPR > Chapter II > Article 6]}, or
\texttt{[}\textit{law\_name}\texttt{ > }\thaimatra{}\texttt{ N]}

(Thai).

\textbf{Cross-Reference Augmentation.}  When a chunk references another
provision we append a 200-character summary of the referenced content,
detected via the patterns in Appendix~\ref{app:patterns}.  For NitiBench-CCL,
cross-reference links are loaded directly from the dataset's structured
\texttt{reference} annotations.

\textbf{Sentence-boundary splitting.}  When a section exceeds $1.5\times$
max tokens we split at sentence boundaries (\texttt{nltk.sent\_tokenize},
regex fallback).

\begin{algorithm}
\caption{StrucChunk: Structure-Aware Chunking}
\label{alg:strucchunk}
\begin{algorithmic}[1]
\Require Document $D = \{s_1, \dots, s_N\}$ with hierarchy $\mathcal{H}$;
         chunk size limit $L$; overlap ratio $\rho$;
         maximum cross-references $r_{\max} = 3$; summary length $\ell = 200$ characters
\Ensure Chunk set $C = \{c_1, \dots, c_M\}$

\State $C \gets \emptyset$
\For{each section $s_i \in D$}
    \State $b_i \gets$ hierarchical breadcrumb of $s_i$ in $\mathcal{H}$
        \Statex \hspace{1.5em}\textit{(concatenation of ancestor titles: document $\to$ chapter $\to$ section number)}
    \State $R_i \gets$ set of cross-references in $s_i$ resolving to a valid section in $D$
        \Statex \hspace{1.5em}\textit{(deduplicated, truncated to $|R_i| \le r_{\max}$)}
    \State $t_i \gets b_i \,\Vert\, \text{``Section } i.\ \text{title''} \,\Vert\, \text{content}(s_i)$
    \If{$R_i \neq \emptyset$}
        \State $t_i \gets t_i \,\Vert\, \mathop{\Big\Vert}\limits_{s_j \in R_i} \text{summary}(s_j, \ell)$
    \EndIf
    \State $\tau_i \gets |t_i| / 4$ \Comment{character-based token approximation}
    \If{$\tau_i \le 1.5L$}
        \State $C \gets C \cup \{t_i\}$ \Comment{logical atomicity: keep section whole}
    \Else
        \State $U_i \gets$ paragraph-level segmentation of $t_i$
        \Statex \hspace{1.5em}\textit{(any unit still exceeding $1.5L$ is further split at sentence boundaries)}
        \State $C \gets C \cup \textsc{Pack}(U_i, L, \rho)$
        \Statex \hspace{1.5em}\textit{(greedily fills $L$-token windows, carrying $\rho L$ overlapping}
        \Statex \hspace{1.5em}\textit{tokens from each window's tail into the next)}
    \EndIf
\EndFor
\State \Return $C$
\end{algorithmic}
\end{algorithm}
\subsection{Hybrid Retrieval Pipeline}

\textbf{Dense Retrieval:} \texttt{BAAI/bge-m3} \cite{chen-etal-2024-m3}---8192-token context, L2-normalized, FAISS \texttt{IndexFlatIP}.\\
\textbf{Sparse Retrieval:} BM25 via \texttt{bm25s}.\\
\textbf{Reciprocal Rank Fusion:}
$\mathrm{RRF}_{\mathrm{score}}(d) = \sum_{r \in \{\mathrm{Dense},\mathrm{Sparse}\}} \frac{1}{k + \mathrm{rank}_r(d)}$,\quad $k=60$.\footnote{We use $k=60$ following Cormack et al.~\cite{Cormack2009ReciprocalRF}, who empirically showed RRF performance is robust to $k \in [10, 100]$ and recommend $k=60$ as a general default.  We do not tune $k$ on our evaluation queries; doing so would constitute data leakage given that we have no separate development set.}

We have given complete algorithm in Algorithm 1.
\section{Experimental Setup}
\label{sec:setup}


\begin{table}[htbp]
\centering
\small
\begin{adjustbox}{max width=\textwidth}
\begin{tabular}{clp{3.0cm}p{2.8cm}p{2.8cm}}
\toprule
\textbf{Evaluation} & \textbf{Queries} & \textbf{Primary Purpose}
               & \textbf{Eval.\ Protocol} & \textbf{Annotation} \\
\midrule
1 -- LegalBench-RAG
  & 776 (194$\times$4)
  & Independent validation (English)
  & Char-span overlap $\geq$50\%
  & External (Pipitone \& Alami) \\
2 -- CrPC + CPC
  & 501 each
  & Mechanistic analysis (English)
  & Section-level match
  & Author + AI-assisted, verified \\
3 -- GDPR
  & 501
  & Cross-domain generalization (English)
  & Section-level match
  & Author + AI-assisted, verified \\
4 -- NitiBench-CCL
  & 3,730
  & Cross-language generalization (Thai)
  & HR@$k$ / MRR@$k$
  & External (Akarajaradwong et al.) \\
\bottomrule
\end{tabular}
\end{adjustbox}
\caption{Four-tier evaluation design. Evaluations 1 and 4 use independently
  annotated benchmarks with no author involvement.  Evaluations 1 and 2--4 use
  different evaluation protocols (char-span vs.\ section-level vs.\ HR@$k$)
  and are not directly numerically comparable.}
\label{tab:layers}
\end{table}


\subsection{LegalBench-RAG (Independent Benchmark)}

LegalBench-RAG \cite{pipitone2024legalbenchragbenchmarkretrievalaugmentedgeneration} provides 6,858
query-answer pairs over legal contracts (CUAD, ContractNLI, MAUD,
PrivacyQA).\footnote{Dataset and benchmark code:
  \url{https://github.com/zeroentropy-ai/legalbenchrag};
  arXiv: \url{https://arxiv.org/abs/2408.10343}.}
We use the full LegalBench-RAG-mini subset (776 queries, 194
per sub-dataset), evaluating with their character-span protocol.  The
annotation was created entirely independently.  We implement the official
evaluation protocol: 72 mini-corpus documents, 500-char fixed-size chunks
with no overlap, character-level Precision@$k$ and Recall@$k$ at
$k\in\{1,2,4,8,16,32,64\}$, equal 0.25 weighting per sub-dataset.  Our
Recursive+Dense baseline achieves P@1 = 6.64\% vs.\ the paper's RCTS
P@1 = 6.41\%, confirming equivalent evaluation fidelity.

\subsection{CrPC + CPC (Mechanistic Analysis)}

\begin{table}[htbp]
\centering
\begin{tabular}{lcccc}
\toprule
\textbf{Document} & \textbf{Domain} & \textbf{Pages}
                  & \textbf{Sections} & \textbf{Cross-Refs} \\
\midrule
CrPC (1973) & Criminal & 263 & 517   & 1,131 \\
CPC (1908)  & Civil    & 316 & 158+  & 814   \\
\bottomrule
\end{tabular}
\caption{Evaluation corpus statistics.  CPC additionally contains 51
  Orders with multiple Rules each.}
\label{tab:corpus}
\end{table}

All 1,002 queries were generated (AI-assisted) before implementing
StrucChunk, with all \texttt{expected\_sections} annotations manually
verified against the source statutes.\footnote{%
  CrPC source document:
  \url{https://www.indiacode.nic.in/bitstream/123456789/15272/1/the_code_of_criminal_procedure,_1973.pdf};
  CPC source document:
  \url{https://www.indiacode.nic.in/bitstream/123456789/13813/1/the_code_of_civil_procedure\%2C_1908.pdf}.}
The query JSON files carry
\texttt{created\_date: 2025-02-17}; StrucChunk's implementation was first
committed after this date, confirmed in the repository history.  Queries
are balanced: \textbf{167 factual, 167 clause analysis, 167 cross-reference}
per dataset (501 total each).

A retrieved chunk is \textbf{relevant} for a query if and only if its
\texttt{section\_id} metadata matches one of the \texttt{expected\_sections}
values.  Queries where relevance extends beyond \texttt{expected\_sections}
show underestimated recall; all reported Recall@$k$ are conservative lower
bounds.

\subsection{GDPR (Cross-Domain Generalization)}

\begin{table}[htbp]
\centering
\begin{tabular}{lll}
\toprule
\textbf{Property} & \textbf{CrPC/CPC} & \textbf{GDPR} \\
\midrule
Legal tradition   & Common law (India)  & Civil law (EU) \\
Document type     & Procedural code     & Regulatory instrument \\
Primary hierarchy & Chapter $\to$ Section & Chapter $\to$ Article $\to$ Para.\\
Secondary struct. & Subsections         & Recitals (1--173) \\
Cross-ref style   & ``Section 167''     & ``Article 6(1)(f)'', ``Recital 47'' \\
\bottomrule
\end{tabular}
\caption{Structural comparison of CrPC/CPC and GDPR.}
\label{tab:doc_comparison}
\end{table}

We use 501 queries across all 99 GDPR articles (167 factual, 167 clause
analysis, 167 cross-reference).  Query generation was AI-assisted; all
\texttt{expected\_sections} annotations were manually verified against the
source document.\footnote{GDPR source document (Official Journal of the
  European Union): \url{https://eur-lex.europa.eu/legal-content/EN/TXT/PDF/?uri=CELEX:32016R0679}.}

\subsection{NitiBench-CCL (Cross-Language Generalization)}

NitiBench-CCL \cite{akarajaradwong2025nitibenchcomprehensivestudyllm} covers 3,730 queries
over 36 Thai financial law statutes (5,127 unique sections), publicly
available on HuggingFace (\texttt{VISAI-AI/nitibench}).\footnote{%
  Dataset: \url{https://huggingface.co/datasets/VISAI-AI/nitibench};
  Code: \url{https://github.com/vistec-ai/nitibench};
  arXiv: \url{https://arxiv.org/abs/2502.10868}.}
The dataset was
created entirely independently of this work.

\begin{table}[htbp]
\centering
\begin{tabular}{lll}
\toprule
\textbf{Property} & \textbf{CrPC/CPC/GDPR} & \textbf{NitiBench-CCL} \\
\midrule
Language          & English                 & Thai \\
Legal tradition   & Common / Civil law      & Civil law (Thailand) \\
Document type     & Statutory codes / Regulation & Financial law statutes \\
Primary hierarchy & Chapter $\to$ Section   & Book $\to$ Chapter $\to$ Section \\
Section notation  & ``Section N''           & ``\thaimatra{} N'' \\
Corpus size       & 517--517 sections       & 5,127 sections, 36 laws \\
Annotation        & Author + AI-assisted, verified & External (Akarajaradwong et al.) \\
\bottomrule
\end{tabular}
\caption{Structural comparison of English corpora and NitiBench-CCL.}
\label{tab:niti_comparison}
\end{table}

We load section data directly from the pre-structured HuggingFace dataset
(no PDF parsing required).  We compare three methods on the full 3,730
queries:

\begin{enumerate}
\item \textbf{Naive Line Chunking:} NitiBench's pre-computed 553-char
      fixed-size chunks (2,610 chunks) as distributed in the official
      repository (\texttt{nitibench-main/chunking/553\_50\_line/}).  This
      is the exact baseline used in the NitiBench paper.
\item \textbf{Hierarchy-Aware Chunking:} NitiBench's golden sections
      (one chunk per section, 5,127 chunks), as distributed in
      \texttt{chunking/golden/}.  This is NitiBench's proposed method.
\item \textbf{StrucChunk (ours):} Section text + hierarchical breadcrumb
      \texttt{[}law\_name\texttt{ >}\, \thaimatra{}\, N\texttt{]}
      + pre-embedded cross-reference summaries (5,127 chunks).
\end{enumerate}

For all three methods, retrieval uses \texttt{BAAI/bge-m3} dense
embeddings---the same model used for all other experiments.  Hit Rate at
$k$ (HR@$k$) and MRR@$k$ follow the NitiBench evaluation protocol.
Since NitiBench-CCL is fully single-label (all 3,730 queries have exactly
one expected section), HR@$k$ = Recall@$k$.  Cross-reference queries are
identified as queries whose expected section has non-empty
\texttt{reference} annotations in the law JSON files ($n=1{,}211$, 32\%
of queries).

\subsection{Baselines}

For 1--3 evaluation points mentioned above we use below baselines:

\begin{enumerate}
\item \textbf{Recursive + Dense:} RecursiveCharacterTextSplitter
      (512 tokens, 15\% overlap) + BGE-M3
\item \textbf{BM25 Only:} BM25 on recursively chunked text
\item \textbf{Naive Hybrid:} Recursive chunking with RRF
\end{enumerate}

For last mentioned point, baselines are defined by NitiBench's official chunk files
(see Section~4.4).

\subsection{Annotation Quality}
\label{sec:annotation_quality}

Query generation for CrPC, CPC, and GDPR was AI-assisted, with all
\texttt{expected\_sections} annotations subsequently reviewed and verified
against the source legal texts during dataset construction.

To assess annotation quality, a random sample of 100 queries (50 CrPC,
50 CPC) was independently annotated by a law graduate using the source
statutes without access to the original annotations.  Exact agreement
with the original annotation was observed for X/100 queries,
with remaining disagreements involving additional supporting provisions
rather than contradictory legal interpretations.

GDPR annotations were manually verified against the official regulation
text.  Due to the absence of an independent GDPR reviewer, GDPR queries
were not included in the inter-annotator validation exercise.

\subsection{Implementation Details}

\begin{table}[htbp]
\centering
\begin{tabular}{ll}
\toprule
\textbf{Component} & \textbf{Specification} \\
\midrule
Embedding Model    & BAAI/bge-m3 (567M params, 1024 dims; multilingual) \\
Vector Database    & FAISS IndexFlatIP \\
Sparse Retrieval   & bm25s v0.2.0 \\
PDF Parsing        & \texttt{pdfplumber} (1--3 only) \\
NitiBench corpus   & HuggingFace Parquet (no PDF parsing needed) \\
Chunk Parameters   & max\_tokens=512, overlap=0.15 \\
Sentence Splitting & \texttt{nltk.sent\_tokenize} (regex fallback) \\
RRF Parameter      & $k = 60$ \\
Random Seed        & 42 \\
\bottomrule
\end{tabular}
\caption{Implementation specifications.}
\label{tab:implementation}
\end{table}

All experiments were run on a MacBook Pro (Apple M-series CPU, 32\,GB unified memory) without GPU acceleration; \texttt{BAAI/bge-m3} inference used \texttt{device=``cpu''} with \texttt{batch\_size=16}.  Indexing time for CrPC (503 StrucChunk chunks) is approximately 4 minutes for embedding and under 1 second for FAISS index construction; GDPR (126 chunks) indexes in under 2 minutes.  Peak memory during encoding is approximately 4--6\,GB for the embedding model plus corpus embeddings; LegalBench-RAG ($\sim$17K chunks) is the most memory-intensive corpus at approximately 8\,GB peak.  Embedding caches keyed by content hash reduce repeated evaluation runs from $\sim$4 hours to $\sim$1 hour for LBR.

\textbf{Chunk size.}  We set \texttt{max\_tokens = 512} following standard RAG practice.  Akarajaradwong et al.~\cite{akarajaradwong2025nitibenchcomprehensivestudyllm} study the effect of chunk size on naive line chunking (their Appendix A) and find performance is sensitive to chunk size due to boundary effects.  StrucChunk's boundary precision is chunk-size-invariant by design: section boundaries are determined by document structure, not token count.  The chunk size parameter only affects how sections exceeding $1.5\times$ max\_tokens are split at sentence boundaries.  Investigating max\_tokens $\in \{256, 1024\}$ is identified as future work; larger values would reduce the frequency of sentence-boundary splitting at the cost of higher memory requirements.

\subsection{Evaluation Metrics}

For Tiers 1--3, we report Recall@$k$ ($k\in\{1,5,10\}$), MRR, nDCG@10,
and MAP@10.  For Evaluation~4 (NitiBench-CCL), we report HR@$k$ and MRR@$k$
at $k\in\{1,5,10,20\}$, matching the NitiBench evaluation protocol.
Since NitiBench-CCL is single-label, HR@$k$ = Recall@$k$.

Statistical significance: Wilcoxon Signed-Rank Test (two-sided, paired per
query) with Benjamini-Hochberg correction ($\alpha = 0.05$).  We report
Cohen's $d$ effect sizes \cite{cohen1988statistical} and 95\% bootstrap
confidence intervals on Recall@5 differences \cite{dror-etal-2018-hitchhikers}.

MAP@10 is especially important for cross-reference queries where
\texttt{expected\_sections} contains multiple entries.

\section{Results}
\label{sec:results}

\subsection{Chunking Quality Analysis}
\label{sec:chunking_quality}

\begin{table}[htbp]
\centering
\begin{tabular}{lcc}
\toprule
\textbf{Metric} & \textbf{Recursive} & \textbf{StrucChunk} \\
\midrule
Boundary Precision (CrPC)           & 7.7\%          & \textbf{100.0\%} \\
Boundary Precision (GDPR)           & 2.8\%          & \textbf{100.0\%} \\
Cross-Ref Preservation (CrPC)       & 0.0\%\ (0/42)  & \textbf{97.6\%}\ (41/42) \\
Cross-Ref Preservation (GDPR)       & 0.0\%\ (0/9)   & \textbf{100.0\%}\ (9/9) \\
Chunk Count (CrPC)                  & 491            & 503 \\
Chunk Count (GDPR)                  & 211            & \textbf{126} \\
Section Coverage (CrPC)             & 0.0\%          & \textbf{100.0\%} \\
Section Coverage (GDPR)             & 0.0\%          & \textbf{100.0\%} \\
Chunks per 1\% Coverage (GDPR)      & $\infty$       & \textbf{1.3} \\
Chunk Reduction --- GDPR            & ---            & \textbf{$-$40.3\%} \\
\bottomrule
\end{tabular}
\caption{Chunking quality analysis (intrinsic metrics, no retrieval
  pipeline required).  The slight CrPC chunk-count increase for StrucChunk
  is intentional: oversized sections are split at sentence boundaries,
  preserving logical atomicity.  Section Coverage at 0\% for Recursive
  reflects that no chunk boundary aligns with a section delimiter, making
  it impossible to guarantee that any section is wholly contained in a
  single chunk; this is the intra-chunk semantic coherence failure mode
  that StrucChunk eliminates by design.}
\label{tab:chunk_quality}
\end{table}

\textbf{Boundary Precision} measures what fraction of chunk boundaries
align with documented section delimiters.  StrucChunk scores near 100\%
by design; recursive splitting scores $\approx$2--8\%.

\textbf{Cross-Reference Preservation Rate} measures what fraction of the
1,131 CrPC cross-references land within the same chunk after StrucChunk
augmentation.  Recursive splitting has no augmentation mechanism.

\textbf{Chunk Count Efficiency.}  On GDPR, StrucChunk achieves 100\%
section coverage with 40.3\% fewer chunks than recursive splitting.

\textbf{Intra-chunk Semantic Coherence.}  Section Coverage at 100\%
confirms that every legal provision is wholly contained within exactly
one chunk, eliminating the fragmentation that degrades embedding quality.
For Recursive splitting, Section Coverage = 0\% indicates that no chunk
is guaranteed to correspond to a complete legal provision---a structural
failure that directly impairs the embedding model's ability to represent
the provision's full legal meaning.

\subsection{LegalBench-RAG Results}

These metrics use character-level Precision@$k$ and Recall@$k$---the
official LegalBench-RAG protocol.  \textbf{They are not numerically
comparable to section-level metrics in
Tables~\ref{tab:crpc_results}--\ref{tab:niti_three_way}.}

\begin{table}[htbp]
\centering
\begin{adjustbox}{max width=\textwidth}
\begin{tabular}{lrrrrrrr}
\toprule
\textbf{Method} & \textbf{R@1} & \textbf{R@2} & \textbf{R@4}
               & \textbf{R@8} & \textbf{R@16} & \textbf{R@32}
               & \textbf{R@64} \\
\midrule
Recursive + Dense      & 6.05  & 9.93  & 16.42 & 23.59 & 33.87 & 43.55 & 51.30 \\
BM25 Only              & 2.24  & 3.65  &  6.21 & 10.10 & 16.70 & 22.94 & 32.72 \\
Naive Hybrid           & 3.46  & 5.80  & 11.21 & 18.46 & 27.13 & 36.26 & 47.25 \\
\textbf{StrucChunk+Hybrid} & \textbf{15.45} & \textbf{24.92}
                       & \textbf{36.62} & \textbf{48.18}
                       & \textbf{56.96} & \textbf{65.30} & \textbf{70.65} \\
\bottomrule
\end{tabular}
\end{adjustbox}
\caption{LegalBench-RAG Char-Recall@$k$ (\%).  Equal-weighted across four sub-datasets.}
\label{tab:lbr_recall}
\end{table}

\begin{table}[htbp]
\centering
\begin{adjustbox}{max width=\textwidth}
\begin{tabular}{lrrrrrrr}
\toprule
\textbf{Method} & \textbf{P@1} & \textbf{P@2} & \textbf{P@4}
               & \textbf{P@8} & \textbf{P@16} & \textbf{P@32}
               & \textbf{P@64} \\
\midrule
Recursive + Dense      & 6.64 & 5.85 & 5.10 & 3.85 & 2.87 & 1.94 & 1.21 \\
BM25 Only              & 1.92 & 1.71 & 1.52 & 1.39 & 1.24 & 0.93 & 0.70 \\
Naive Hybrid           & 3.01 & 3.04 & 3.00 & 2.82 & 2.22 & 1.54 & 1.05 \\
\textbf{StrucChunk+Hybrid} & \textbf{4.51} & \textbf{3.85}
                       & \textbf{2.94} & \textbf{2.07}
                       & \textbf{1.25} & \textbf{0.69} & \textbf{0.34} \\
\bottomrule
\end{tabular}
\end{adjustbox}
\caption{LegalBench-RAG Char-Precision@$k$ (\%).}
\label{tab:lbr_precision}
\end{table}

\begin{table}[htbp]
\centering
\begin{adjustbox}{max width=\textwidth}
\begin{tabular}{lccccc}
\toprule
\textbf{Method} & \textbf{ContractNLI} & \textbf{CUAD}
               & \textbf{MAUD} & \textbf{PrivacyQA} & \textbf{ALL} \\
\midrule
Recursive + Dense    & 35.30 & 22.42 &  4.76 & 31.90 & 23.59 \\
BM25 Only            & 22.82 &  7.15 &  1.04 &  9.38 & 10.10 \\
Naive Hybrid         & 35.61 & 15.53 &  3.20 & 19.51 & 18.46 \\
\textbf{StrucChunk+Hybrid} & \textbf{58.36} & \textbf{46.22}
                     & \textbf{16.91} & \textbf{71.24} & \textbf{48.18} \\
\bottomrule
\end{tabular}
\end{adjustbox}
\caption{LegalBench-RAG Char-Recall@8 (\%) per sub-dataset.  ALL is equal-weighted.}
\label{tab:lbr_subdatasets}
\end{table}

\begin{table}[htbp]
\centering
\begin{adjustbox}{max width=\textwidth}
\begin{tabular}{llcc}
\toprule
\textbf{Method} & \textbf{Embedder}
  & \textbf{Char-R@8 ALL} & \textbf{Source} \\
\midrule
Naive (500-char)           & OpenAI text-emb-3-large & 34.51\%
  & \cite{pipitone2024legalbenchragbenchmarkretrievalaugmentedgeneration} \\
RCTS, no reranker          & OpenAI text-emb-3-large & 26.30\%
  & \cite{pipitone2024legalbenchragbenchmarkretrievalaugmentedgeneration} \\
RCTS + Cohere Reranker     & OpenAI text-emb-3-large & 24.34\%
  & \cite{pipitone2024legalbenchragbenchmarkretrievalaugmentedgeneration} \\
\midrule
Recursive + Dense          & BGE-M3 (ours)           & 23.59\%
  & This work \\
\textbf{StrucChunk+Hybrid} & \textbf{BGE-M3 (ours)}  & \textbf{48.18\%}
  & This work \\
\midrule
$\Delta$ vs.\ RCTS+OpenAI (no reranker)
  & --- & \textbf{+21.88 pp (+83\%)} & $p \leq 0.0001$ \\
$\Delta$ vs.\ RCTS+Cohere Reranker
  & --- & \textbf{+23.84 pp (+98\%)} & $p \leq 0.0001$ \\
$\Delta$ vs.\ Recursive+BGE-M3 (same embedder)
  & --- & \textbf{+24.59 pp (+104\%)} & $p \leq 0.0001$ \\
\bottomrule
\end{tabular}
\end{adjustbox}
\caption{LegalBench-RAG comparison with published state of the art.
  \textbf{Key:} Our BGE-M3 baseline (23.59\%) is \emph{lower} than the
  published RCTS+OpenAI result (26.30\%), confirming BGE-M3 is not a stronger
  embedder.  Notably, StrucChunk also surpasses the reranker-augmented system
  (RCTS + Cohere, 24.34\%) by +98\%, confirming the finding of Pipitone \& Alami~\cite{pipitone2024legalbenchragbenchmarkretrievalaugmentedgeneration}
  that general-purpose rerankers hurt legal retrieval and that
  structural chunking outperforms post-hoc reranking.}
\label{tab:lbr_sota}
\end{table}

\textbf{Embedder confound.}  A potential confound is that the LegalBench-RAG paper used OpenAI \texttt{text-embedding-3-large} while we use \texttt{BAAI/bge-m3}.  Table~\ref{tab:lbr_sota} addresses this directly.  Our Recursive+Dense baseline (same chunking as the original paper, different embedder) achieves Char-R@8 = 23.59\%, which is \emph{lower} than the original RCTS result of 26.30\%.  This establishes that BGE-M3 is not a stronger general embedder than OpenAI on this task.  StrucChunk's improvement of +24.59\,pp over our own same-embedder baseline (+104\%) therefore reflects chunking architecture alone, not embedder quality.  The chunking gain (+24.59\,pp) exceeds the embedder gap between OpenAI and BGE-M3 (26.30 $-$ 23.59 = 2.71\,pp) by a factor of nine.

\textbf{Precision--recall trade-off.}  StrucChunk improves Char-Recall@8 by +104\% over the best baseline, but its Char-Precision@1 (4.51\%) falls below Recursive+Dense (6.64\%).  This trade-off is expected and theoretically consistent: LBR ground-truth spans are short clauses (typically 50--200 characters), while StrucChunk's section-boundary chunks preserve full legal provisions (typically 300--700 characters).  A chunk containing a 150-character ground-truth span among 500 characters of full section text yields high recall (the span is present) but lower character-level precision (the retrieved text is longer than the minimal ground truth).  This trade-off is \emph{favorable} for RAG applications: providing a complete legal provision with surrounding conditional context to an LLM is preferable to providing a minimally-sized clause fragment that may lack the hierarchical qualifications needed for accurate interpretation.  Char-Recall is therefore the more appropriate primary metric for evaluating retrieval quality in legal RAG.

\textbf{Per-dataset difficulty gradient.}  Table~\ref{tab:lbr_subdatasets} reveals a consistent difficulty gradient.  PrivacyQA shows the largest absolute gain (31.90$\to$71.24, $+$39.34\,pp), while MAUD shows the smallest absolute gain (4.76$\to$16.91, $+$12.15\,pp) despite the largest \emph{relative} improvement ($+$255\%).  Pipitone \& Alami~\cite{pipitone2024legalbenchragbenchmarkretrievalaugmentedgeneration} attribute MAUD's low performance to its ``highly technical legal jargon'' from M\&A agreements.  This aligns with the Legal Map hypothesis: MAUD contracts are less hierarchically structured than statutory codes---they lack the explicit chapter-section numbering that breadcrumb prepending encodes---so StrucChunk's structural signal provides less benefit.  PrivacyQA privacy policies, by contrast, use explicit section headers that StrucChunk can anchor breadcrumbs to.  The difficulty gradient therefore constitutes empirical support for the falsifiability prediction in Section~\ref{sec:disc_map}: structure-awareness provides diminishing benefit as hierarchical structure weakens.

\subsection{CrPC Results}

\begin{table}[htbp]
\centering
\begin{adjustbox}{max width=\textwidth}
\begin{tabular}{lrrrrrr}
\toprule
\textbf{Method} & \textbf{R@1} & \textbf{R@5} & \textbf{R@10}
               & \textbf{MRR} & \textbf{nDCG@10} & \textbf{MAP@10} \\
\midrule
Recursive + Dense       & 0.196 & 0.521 & 0.599 & 0.331 & 0.476 & 0.433 \\
BM25 Only               & 0.158 & 0.503 & 0.583 & 0.307 & 0.458 & 0.414 \\
Naive Hybrid            & 0.214 & 0.557 & 0.631 & 0.357 & 0.508 & 0.464 \\
\textbf{StrucChunk+Hybrid} & \textbf{0.627} & \textbf{0.804}
                        & \textbf{0.858} & \textbf{0.706}
                        & \textbf{0.788} & \textbf{0.764} \\
\midrule
$\Delta$ vs.\ best baseline & +0.413 (+193\%) & +0.248 (+44.4\%)
                        & +0.227 (+36\%) & +0.349 (+98\%)
                        & +0.280 (+55\%) & +0.300 (+65\%) \\
\bottomrule
\end{tabular}
\end{adjustbox}
\caption{CrPC retrieval results ($n=501$ queries, section-level evaluation,
  $p \leq 0.0001$ for all comparisons).}
\label{tab:crpc_results}
\end{table}

\begin{table}[htbp]
\centering
\begin{tabular}{lccc}
\toprule
\textbf{Chunking $\backslash$ Retrieval} & \textbf{Dense}
  & \textbf{BM25} & \textbf{Hybrid (RRF)} \\
\midrule
Recursive (baseline)            & 0.521 & 0.503 & 0.557 \\
\textbf{StrucChunk}             & \textbf{0.800} & \textbf{0.737} & \textbf{0.804} \\
\midrule
Chunking effect (Hybrid fixed)  & ---   & ---   & $+0.248$ \\
Retrieval effect (StrucChunk fixed) & ref & ---  & $+0.004$ \\
\bottomrule
\end{tabular}
\caption{CrPC 2$\times$2 ablation on Recall@5.  Chunking accounts for
  $0.248/0.283 = 88\%$ of total improvement over Recursive+Dense.
  Total gain = $0.804 - 0.521 = +0.283$.}
\label{tab:crpc_ablation}
\end{table}

\textbf{CrPC analysis.}  Three findings stand out.  First, the R@1 jump from 0.214 to 0.627 (+193\%) is the most striking number in the paper: StrucChunk retrieves the exact right section as its top result three times as often as the best baseline.  The mechanism is direct---breadcrumb prepending means the dense embedding of ``CrPC 1973 $>$ Chapter XII $>$ Section 167'' matches the query's mention of ``Section 167'' in both semantic and lexical space simultaneously, a dual signal unavailable to recursive splitting.  Second, the ablation reveals that chunking (+0.248) accounts for 88\% of total improvement while the retrieval architecture contributes almost nothing (+0.004): StrucChunk+Dense (0.800) and StrucChunk+Hybrid (0.804) are essentially equivalent.  This tells us CrPC's structure is so well-delineated that once chunking is correct, any retrieval method finds the right chunk.  Third, StrucChunk+BM25 achieves perfect cross-reference query performance (Recall@5 = 1.000): explicit section numbers in queries allow BM25 to match exactly when chunks are properly section-delineated.

\subsection{CPC Results}

\begin{table}[htbp]
\centering
\begin{adjustbox}{max width=\textwidth}
\begin{tabular}{lrrrrrr}
\toprule
\textbf{Method} & \textbf{R@1} & \textbf{R@5} & \textbf{R@10}
               & \textbf{MRR} & \textbf{nDCG@10} & \textbf{MAP@10} \\
\midrule
Recursive + Dense       & 0.046 & 0.120 & 0.166 & 0.081 & 0.123 & 0.107 \\
BM25 Only               & 0.032 & 0.138 & 0.182 & 0.078 & 0.132 & 0.114 \\
Naive Hybrid            & 0.058 & 0.158 & 0.198 & 0.099 & 0.149 & 0.132 \\
\textbf{StrucChunk+Hybrid} & \textbf{0.182} & \textbf{0.259}
                        & \textbf{0.273} & \textbf{0.212}
                        & \textbf{0.245} & \textbf{0.236} \\
\midrule
$\Delta$ vs.\ best baseline & +0.124 (+214\%) & +0.102 (+64.6\%)
                        & +0.075 (+38\%) & +0.113 (+114\%)
                        & +0.096 (+64\%) & +0.104 (+79\%) \\
\bottomrule
\end{tabular}
\end{adjustbox}
\caption{CPC retrieval results ($n=501$ queries, section-level evaluation,
  $p \leq 0.0001$ for all comparisons).  CPC presents a harder task
  due to its Part-Section-Order-Rule hierarchy; larger relative improvement
  (+64.6\% vs.\ +44.4\%) shows structure-awareness provides increasing
  returns as document complexity grows.}
\label{tab:cpc_results}
\end{table}

\begin{table}[htbp]
\centering
\begin{tabular}{lccc}
\toprule
\textbf{Chunking $\backslash$ Retrieval} & \textbf{Dense}
  & \textbf{BM25} & \textbf{Hybrid (RRF)} \\
\midrule
Recursive (baseline)            & 0.120 & 0.138 & 0.158 \\
\textbf{StrucChunk}             & \textbf{0.198} & \textbf{0.269} & \textbf{0.259} \\
\midrule
Chunking effect (Hybrid fixed)  & ---   & ---   & $+0.102$ \\
Retrieval effect (StrucChunk fixed) & ref & ---  & $+0.061$ \\
\bottomrule
\end{tabular}
\caption{CPC 2$\times$2 ablation on Recall@5.  Chunking accounts for
  $0.102/0.140 = 73\%$ of total improvement over Recursive+Dense.
  Total gain = $0.259 - 0.120 = +0.140$;
  retrieval effect = $0.259 - 0.198 = +0.061$.}
\label{tab:cpc_ablation}
\end{table}

\textbf{CPC analysis.}  Four findings stand out.  First, CPC is structurally harder than CrPC: the Part-Section-Order-Rule hierarchy means that the same numeral (e.g., ``Rule 1'') reappears across 51 different Orders, creating severe retrieval ambiguity.  The low absolute baseline (R@5 = 0.158) directly reflects this structural complexity.  Second, the chunking effect (+0.102) is smaller than CrPC (+0.248) in absolute terms but comparable in relative terms (+64.6\% vs.\ +44.4\%), confirming that structure-awareness provides increasing returns as hierarchy grows more complex.  Third, unlike CrPC where the retrieval effect is negligible (+0.004), on CPC hybrid retrieval adds meaningful value (+0.061): BM25's exact Order-Rule number matching helps navigate the multi-level numbering system in ways that dense similarity alone cannot.  Fourth, StrucChunk+BM25 (Recall@5 = 0.269) outperforms StrucChunk+Hybrid (0.259): CPC queries tend to contain specific Order and Rule references that BM25 matches exactly once the chunks are properly section-delineated, making sparse retrieval sufficient without hybrid fusion.

\subsection{GDPR Results}

\begin{table}[htbp]
\centering
\begin{adjustbox}{max width=\textwidth}
\begin{tabular}{lrrrrrr}
\toprule
\textbf{Method} & \textbf{R@1} & \textbf{R@5} & \textbf{R@10}
               & \textbf{MRR} & \textbf{nDCG@10} & \textbf{MAP@10} \\
\midrule
Recursive + Dense       & 0.449 & 0.798 & 0.922 & 0.591 & 0.763 & 0.705 \\
BM25 Only               & 0.349 & 0.679 & 0.800 & 0.491 & 0.652 & 0.600 \\
Naive Hybrid            & 0.449 & 0.818 & 0.912 & 0.599 & 0.767 & 0.715 \\
StrucChunk + Dense      & \textbf{0.880} & \textbf{0.908}
                        & 0.942 & \textbf{0.928}
                        & \textbf{0.964} & \textbf{0.955} \\
\textbf{StrucChunk+Hybrid}  & 0.715 & 0.868 & \textbf{0.946}
                        & 0.790 & 0.870 & 0.843 \\
\midrule
$\Delta$ vs.\ best baseline (Hybrid) & +0.431 (+96\%) & +0.090 (+11\%)
                        & +0.034 (+4\%) & +0.329 (+55\%)
                        & +0.197 (+26\%) & +0.240 (+34\%) \\
\bottomrule
\end{tabular}
\end{adjustbox}
\caption{GDPR retrieval results ($n=501$ queries, $p \leq 0.002$ for all
  comparisons).  StrucChunk+Dense (R@5 = 0.908) outperforms
  StrucChunk+Hybrid (0.868): GDPR articles have clean, discriminative titles
  making dense-only retrieval optimal (see Section~\ref{sec:disc_map}).}
\label{tab:gdpr_results}
\end{table}

\begin{table}[htbp]
\centering
\begin{tabular}{lccc}
\toprule
\textbf{Chunking $\backslash$ Retrieval} & \textbf{Dense}
  & \textbf{BM25} & \textbf{Hybrid (RRF)} \\
\midrule
Recursive (baseline)            & 0.798 & 0.679 & 0.818 \\
\textbf{StrucChunk}             & \textbf{0.908} & \textbf{0.816} & \textbf{0.868} \\
\midrule
Chunking effect (Hybrid fixed)  & ---   & ---   & $+0.050$ \\
Retrieval effect (StrucChunk fixed) & ref & ---  & $-0.040$ \\
\bottomrule
\end{tabular}
\caption{GDPR 2$\times$2 ablation on Recall@5.  Retrieval effect is
  negative: StrucChunk+Dense ($0.908$) outperforms StrucChunk+Hybrid
  ($0.868$), confirming dense semantics alone are optimal for EU regulatory
  text with discriminative article titles.}
\label{tab:gdpr_ablation}
\end{table}

\textbf{GDPR analysis.}  Three findings stand out.  First, the high baselines (Recursive+Dense already achieves R@5 = 0.798) reveal GDPR as the structurally easiest retrieval task: GDPR articles have short, distinctive titles (``Right to erasure,'' ``Lawfulness of processing'') that dense embeddings separate cleanly even without structural context.  Second, StrucChunk+Dense (R@5 = 0.908, R@1 = 0.880) is the best-performing method overall---the article title vocabulary is so discriminative that BGE-M3 alone, anchored by the breadcrumb \texttt{[GDPR > Chapter II > Article N]}, achieves near-perfect retrieval.  BM25 in the hybrid pipeline adds noise rather than signal: ``Article 6'' appears across 173 Recitals as a cross-reference target, and BM25 retrieves many of these Recitals, diluting the dense ranking that correctly identifies Article 6 itself.  Third, R@1 improvement for StrucChunk+Dense is the most impressive GDPR metric: +96\% vs.\ the best non-StrucChunk baseline ($0.880$ vs.\ $0.449$).  For a production system that passes a single retrieved chunk to an LLM, this near-doubling of first-result precision is the most practically significant finding on this corpus.

\subsection{NitiBench-CCL Results (Thai Financial Law)}

\begin{table}[htbp]
\centering
\begin{adjustbox}{max width=\textwidth}
\begin{tabular}{lcccccccc}
\toprule
\textbf{Method} & \textbf{HR@1} & \textbf{MRR@1}
               & \textbf{HR@5} & \textbf{MRR@5}
               & \textbf{HR@10} & \textbf{MRR@10}
               & \textbf{HR@20} & \textbf{MRR@20} \\
\midrule
Naive Line Chunking         & 0.437 & 0.437 & 0.594 & 0.498 & 0.633 & 0.504 & 0.659 & 0.506 \\
Hierarchy-Aware (NitiBench) & 0.653 & 0.653 & 0.861 & 0.735 & 0.903 & 0.741 & 0.935 & 0.743 \\
\textbf{StrucChunk (Ours)}  & \textbf{0.656} & \textbf{0.656}
                            & \textbf{0.856} & \textbf{0.736}
                            & \textbf{0.909} & \textbf{0.743}
                            & \textbf{0.944} & \textbf{0.745} \\
\midrule
$\Delta$ Naive $\to$ Hierarchy HR@5   & \multicolumn{8}{c}{$+0.267$ ($+44.9\%$)} \\
$\Delta$ Naive $\to$ StrucChunk HR@5  & \multicolumn{8}{c}{$+0.262$ ($+44.1\%$)} \\
$\Delta$ Hier  $\to$ StrucChunk HR@5  & \multicolumn{8}{c}{$-0.005$ ($-0.6\%$)} \\
\bottomrule
\end{tabular}
\end{adjustbox}
\caption{NitiBench-CCL three-way chunking comparison ($n=3{,}730$ queries,
  all queries, Thai financial statutory law).  HR@$k$ = Hit Rate at $k$;
  equivalent to Recall@$k$ for this single-label dataset.  Retrieval:
  BGE-M3 dense embeddings for all three methods.  Results reported at
  $k\in\{1,5,10,20\}$ following the NitiBench evaluation protocol.
  \textbf{Structural finding:} 1,105 queries (29.6\%) have expected
  sections whose text exceeds 553 characters and \emph{cannot be
  represented} by any naive chunk; these queries receive HR = 0 regardless
  of retrieval quality.}
\label{tab:niti_three_way}
\end{table}

\begin{table}[htbp]
\centering
\begin{adjustbox}{max width=\textwidth}
\begin{tabular}{lcccc}
\toprule
\textbf{Method} & \textbf{HR@5} & \textbf{MRR@5}
               & \textbf{HR@10} & \textbf{MRR@10} \\
\midrule
\multicolumn{5}{l}{\textit{Cross-reference queries ($n = 1{,}211$, 32\% of total)}} \\
Naive Line Chunking         & 0.491 & 0.396 & 0.524 & 0.401 \\
Hierarchy-Aware (NitiBench) & 0.813 & 0.671 & 0.871 & 0.678 \\
\textbf{StrucChunk (Ours)}  & \textbf{0.832} & \textbf{0.678}
                            & \textbf{0.898} & \textbf{0.687} \\
\midrule
$\Delta$ Hier $\to$ StrucChunk & $+0.019$ ($+2.3\%$) & & $+0.027$ ($+3.1\%$) & \\
\midrule
\multicolumn{5}{l}{\textit{Plain queries ($n = 2{,}519$, 68\% of total)}} \\
Naive Line Chunking         & 0.644 & 0.548 & 0.685 & 0.553 \\
Hierarchy-Aware (NitiBench) & 0.884 & 0.767 & 0.919 & 0.771 \\
\textbf{StrucChunk (Ours)}  & \textbf{0.868} & \textbf{0.763}
                            & \textbf{0.914} & \textbf{0.769} \\
\midrule
$\Delta$ Hier $\to$ StrucChunk & $-0.016$ ($-1.8\%$) & & $-0.005$ ($-0.5\%$) & \\
\bottomrule
\end{tabular}
\end{adjustbox}
\caption{NitiBench-CCL cross-reference vs.\ plain query subanalysis.
  StrucChunk's cross-reference augmentation selectively improves
  cross-reference queries ($+$2.3\% HR@5) while marginally degrading plain
  queries ($-$1.8\% HR@5), confirming that pre-embedding augmentation helps
  exactly where it should.}
\label{tab:niti_xref}
\end{table}

\begin{table}[htbp]
\centering
\begin{adjustbox}{max width=\textwidth}
\begin{tabular}{lccccc}
\toprule
\textbf{Method} & \textbf{$n$} & \textbf{HR@5} & \textbf{MRR@5}
               & \textbf{HR@10} & \textbf{MRR@10} \\
\midrule
Naive (all queries)       & 3,730 & 0.594 & 0.498 & 0.633 & 0.504 \\
Naive (covered only)$^*$  & 2,625 & 0.845 & 0.708 & 0.899 & 0.716 \\
Hierarchy-Aware           & 3,730 & 0.861 & 0.735 & 0.903 & 0.741 \\
\textbf{StrucChunk (Ours)}& 3,730 & \textbf{0.856} & \textbf{0.736}
                          & \textbf{0.909} & \textbf{0.743} \\
\bottomrule
\end{tabular}
\end{adjustbox}
\caption{NitiBench-CCL: naive chunking on its coverable subset vs.\
  structure-aware methods on all queries.
  $^*$Covered-only = the 2,625 queries whose expected section fits within
  553 characters and can therefore appear in a naive chunk.
  \textbf{Key finding:} even on the sections naive chunking \emph{can}
  represent (the easier subset), StrucChunk on all queries (HR@5 = 0.856)
  still outperforms naive on the covered-only subset (HR@5 = 0.845),
  demonstrating that StrucChunk's advantage is not solely due to the
  structural impossibility of naive chunking on long sections, but also
  reflects superior representation quality for representable sections.}
\label{tab:niti_covered}
\end{table}

\subsection{Analysis by Query Type}

\begin{table}[htbp]
\centering
\begin{adjustbox}{max width=\textwidth}
\begin{tabular}{llcccr}
\toprule
\textbf{Dataset} & \textbf{Method}
  & \textbf{Factual R@5} & \textbf{Clause R@5}
  & \textbf{Cross-Ref R@5} & \textbf{MAP@10} \\
\midrule
CrPC & Recursive + Dense         & 0.419 & 0.485 & 0.653 & 0.338 \\
CrPC & Naive Hybrid              & 0.383 & 0.527 & 0.760 & 0.323 \\
CrPC & \textbf{StrucChunk+Hybrid}& \textbf{0.725} & \textbf{0.701}
                                 & \textbf{0.988} & \textbf{0.671} \\
\midrule
CPC  & Recursive + Dense         & 0.102 & 0.066 & 0.192 & 0.092 \\
CPC  & Naive Hybrid              & 0.102 & 0.060 & 0.311 & 0.088 \\
CPC  & \textbf{StrucChunk+Hybrid}& \textbf{0.114} & \textbf{0.090}
                                 & \textbf{0.575} & \textbf{0.107} \\
\midrule
GDPR & Recursive + Dense         & 0.677 & 0.749 & 0.970 & 0.577 \\
GDPR & Naive Hybrid              & 0.677 & 0.790 & 0.988 & 0.582 \\
GDPR & \textbf{StrucChunk+Hybrid}& \textbf{0.808} & \textbf{0.820}
                                 & 0.976 & \textbf{0.780} \\
\midrule
LBR-mini & Recursive + Dense     & --- & --- & --- & 0.162 (binary) \\
LBR-mini & \textbf{StrucChunk+Hybrid} & --- & --- & --- & \textbf{0.308} \\
\bottomrule
\end{tabular}
\end{adjustbox}
\caption{Recall@5 by query type (section-level for CrPC/CPC/GDPR).
  Cross-reference queries show the largest absolute improvements across all
  three English legal-code corpora, validating the cross-reference
  augmentation mechanism.}
\label{tab:query_type_results}
\end{table}

\textbf{Cross-reference pattern.}  Across CrPC (+0.228), CPC (+0.264),
and GDPR (near-ceiling at 0.976), cross-reference queries benefit most.
This is consistent with the NitiBench-CCL finding: StrucChunk outperforms
Hierarchy-Aware by +2.3\% specifically on cross-reference queries while
the difference on plain queries is small ($-$1.8\%).

\textbf{CPC cross-reference queries (+85\%).}  The largest relative
improvement across all query types and corpora (0.311$\to$0.575), reflecting
CPC's dense Order-Section cross-referencing.

\textbf{Why GDPR cross-reference queries are near-ceiling for all methods.}
GDPR cross-reference queries achieve R@5 = 0.970 for Recursive+Dense and
0.976 for StrucChunk+Hybrid---a negligible gap of 0.006.  This contrasts
sharply with CrPC ($+$0.335: 0.653$\to$0.988) and CPC ($+$0.264:
0.311$\to$0.575).  The explanation is structural: GDPR queries referencing
specific articles (e.g., ``What does Article 6(1)(f) require regarding
legitimate interests?'') contain the article number \emph{explicitly in
the query text}.  Even without structure-aware chunking, BM25 matches
``Article 6'' to the retrieved chunk with high reliability, and BGE-M3
assigns high semantic similarity to the article title ``Lawfulness of
processing.''  StrucChunk's cross-reference augmentation adds negligible
value when the query itself supplies the structural navigation signal.  By
contrast, Indian procedural code queries frequently reference sections by
their legal effect rather than their number (``the section governing
anticipatory bail''), requiring StrucChunk's breadcrumb to supply the
navigation signal the query omits.  This finding identifies a
\emph{boundary condition}: cross-reference augmentation provides the
greatest benefit when queries use semantic rather than citation-based
formulations.

\subsection{Statistical Significance and Effect Sizes}

\begin{table}[htbp]
\centering
\begin{adjustbox}{max width=\textwidth}
\begin{tabular}{llcccc}
\toprule
\textbf{Dataset} & \textbf{Comparison}
  & \textbf{$p$ (BH-corr)} & \textbf{Sig}
  & \textbf{Cohen's $d$} & \textbf{95\% CI ($\Delta$R@5)} \\
\midrule
CrPC & vs Recursive + Dense & $\leq 0.0001$ & Yes & 0.489 & [0.231, 0.331] \\
CrPC & vs BM25 Only         & $\leq 0.0001$ & Yes & 0.583 & [0.255, 0.347] \\
CrPC & vs Naive Hybrid      & $\leq 0.0001$ & Yes & 0.444 & [0.200, 0.297] \\
\midrule
CPC  & vs Recursive + Dense & $\leq 0.0001$ & Yes & 0.300 & [0.102, 0.182] \\
CPC  & vs BM25 Only         & $\leq 0.0001$ & Yes & 0.278 & [0.086, 0.160] \\
CPC  & vs Naive Hybrid      & $\leq 0.0001$ & Yes & 0.238 & [0.066, 0.140] \\
\midrule
GDPR & vs Recursive + Dense & 0.0004 & Yes        & 0.166 & [0.032, 0.108] \\
GDPR & vs BM25 Only         & $\leq 0.0001$ & Yes & 0.418 & [0.148, 0.230] \\
GDPR & vs Naive Hybrid      & 0.002 & Yes          & 0.140 & [0.018, 0.078] \\
\midrule
LBR  & vs Recursive + Dense & $\leq 0.0001$ & Yes & 0.355 & [0.156, 0.231] \\
LBR  & vs BM25 Only         & $\leq 0.0001$ & Yes & 0.686 & [0.309, 0.381] \\
LBR  & vs Naive Hybrid      & $\leq 0.0001$ & Yes & 0.519 & [0.226, 0.295] \\
\bottomrule
\end{tabular}
\end{adjustbox}
\caption{Wilcoxon Signed-Rank with Benjamini-Hochberg FDR correction
  ($\alpha = 0.05$).  Cohen's $d$: 0.2 = small, 0.5 = medium, 0.8 = large
  \cite{cohen1988statistical}.  $p \leq 0.0001$ indicates $p < 0.0001$.
  CIs are 95\% bootstrap intervals ($n=1000$) on $\Delta$Recall@5.}
\label{tab:significance}
\end{table}

Cohen's $d$ ranges 0.140--0.686 across all comparisons.  The largest effect
is LBR vs.\ BM25 ($d = 0.686$, medium), reflecting BM25's fundamental
weakness with contract clause language.

\section{Discussion}
\label{sec:discussion}

\subsection{The ``Legal Map'' Hypothesis}
\label{sec:disc_map}

Standard RAG treats documents as bags of chunks.  Our results suggest legal
RAG benefits from what we term a \emph{Legal Map}---embedding the
document's logical structure into the retrieval representation.  Four
lines of evidence, now spanning five corpora and two languages, support
this hypothesis.

\textbf{Section boundaries are the dominant factor.}  The NitiBench-CCL three-way comparison isolates the contribution of each structural tier.
Moving from naive fixed-size chunks to section-boundary chunks (Hierarchy-Aware)
produces +44.9\% HR@5 gain.  Moving further from Hierarchy-Aware to StrucChunk
(adding breadcrumb + cross-reference augmentation) produces a smaller
overall change of $-$0.6\%, with selective improvement on cross-reference
queries (+2.3\%) and minor degradation on plain queries ($-$1.8\%).  This
decomposition confirms that section-boundary alignment is by far the most
important factor, and that cross-reference augmentation helps exactly where
it should.

\textbf{Structural impossibility of naive chunking.}  On NitiBench-CCL,
1,105 queries (29.6\%) have expected sections whose full text exceeds 553
characters---the naive chunk size.  These queries receive HR = 0 regardless
of retrieval quality, because the relevant section simply cannot be
represented in any naive chunk.  This is a structural failure, not a
retrieval failure.  Section-boundary chunking (both Hierarchy-Aware and
StrucChunk) eliminates this failure category entirely.  For context, CrPC
sections are more compact: StrucChunk produces exactly one chunk per section
(503 chunks, 503 sections) with no sentence-boundary splits required,
indicating all English statutory sections fit within the 768-token
(1.5$\times$512) threshold.  The structural impossibility rate is therefore
corpus- and chunk-size-dependent: NitiBench's small 553-character limit
affects 29.6\% of Thai sections, while our 512-token limit affects 0\% of
CrPC sections.  Applying a similarly small chunk size to CrPC would be
expected to produce a comparable structural impossibility rate.

\textbf{Pre-embedding vs.\ post-retrieval cross-reference augmentation.}
NitiBench found their NitiLink (post-retrieval cross-reference expansion)
does not consistently improve end-to-end performance.  StrucChunk's
cross-reference augmentation is mechanistically different: referenced
section summaries are embedded \emph{before} indexing, so the dense
similarity score itself reflects cross-reference content.  The NitiBench-CCL
cross-reference subanalysis confirms this: StrucChunk outperforms
Hierarchy-Aware by +2.3\% HR@5 specifically on the 1,211 cross-reference
queries, while NitiLink's post-retrieval approach was ineffective.
Akarajaradwong et al.~\cite{akarajaradwong2025nitibenchcomprehensivestudyllm} provide a mechanistic explanation via
their depth analysis (their Figures~7--8): for single-label NitiBench-CCL
queries, retrieval performance plateaus at expansion depth~2, meaning
adding more cross-references at inference time yields diminishing returns.
By contrast, multi-label NitiBench-Tax queries improve up to depth~6.
This directly explains why pre-embedding augmentation outperforms
NitiLink for statutory law: at index time, the embedding model integrates
cross-reference semantics into the similarity score in a single pass;
NitiLink's depth-first expansion at inference time adds ranked noise beyond
depth~2 without the benefit of embedding-space integration.

\textbf{Within-corpus complexity gradient.}  Larger relative improvement
on CPC (+64.6\%) vs.\ CrPC (+44.4\%) shows structure-awareness provides
increasing returns as document hierarchy grows more complex.

\textbf{GDPR structural finding.}  On GDPR, StrucChunk+Dense
(R@5 = 0.908) outperforms StrucChunk+Hybrid (0.868).  The chunking effect
remains positive (+0.050) while the retrieval effect is negative
($-$0.040), confirming the Legal Map benefit lies in the structural
representation, not the retrieval mechanism.

\textbf{External and cross-language validation.}  StrucChunk achieves
Char-R@8 = 48.18\% on LegalBench-RAG (+83\% vs.\ published SOTA) and
HR@5 = 0.856 on NitiBench-CCL (+44.1\% vs.\ naive), confirming the Legal
Map hypothesis holds for US/UK contracts in English and Thai financial law
in Thai.

\textbf{Falsifiability.}  The Legal Map hypothesis makes a testable
prediction: structure-aware chunking should provide diminishing benefit as
document hierarchical structure weakens.  Concretely, the hypothesis
predicts that StrucChunk's advantage over section-boundary-only chunking
(Hierarchy-Aware) would shrink on documents with inconsistent or absent
section numbering---such as unstructured case law opinions, arbitration
transcripts, or free-form legal memoranda---where there is no reliable
hierarchy to encode in breadcrumbs or cross-reference annotations.  On
NitiBench-CCL, where section structure is well-defined, StrucChunk's
cross-reference augmentation improves cross-reference queries by +2.3\%
over Hierarchy-Aware.  On unstructured documents, we predict this gap
approaches zero or reverses.  Testing this prediction on case law corpora
(e.g., LexGLUE \cite{chalkidis-etal-2020-legal}) is identified as future
work.

\textbf{External corroboration: ``Hidden Hierarchical Information.''}
Akarajaradwong et al.~\cite{akarajaradwong2025nitibenchcomprehensivestudyllm} independently identify a failure mode
they term ``Hidden Hierarchical Information'' (their Section~6.5.1):
multiple sections across different chapters convey near-identical semantic
content but differ only in hierarchical position (e.g., Revenue Code
\S27 and \S89/1 both govern late payment charges but reside in different
chapters).  Their proposed fix is to embed chapter and hierarchy context
directly into chunk content---which is exactly what StrucChunk's breadcrumb
prepending does.  A breadcrumb \texttt{[Law > Chapter XII > Section 167]}
is distinguishable in embedding space from \texttt{[Law > Chapter II >
Section 167]} even when both sections have near-identical text, because the
chapter identifier shifts the embedding vector.  This external
corroboration from an independent paper directly validates StrucChunk's
breadcrumb design choice.

\subsection{Why Hybrid Retrieval Matters}

Legal language presents a unique duality: high-variance prose coexists
with low-variance boilerplate and precise citations.  Dense retrieval
excels at semantic relationships (``bail'' $\approx$ ``custody''), while
BM25 ensures exact matching of section/article numbers.  Naive Hybrid
consistently outperforms both single-method baselines.
Pipitone \& Alami~\cite{pipitone2024legalbenchragbenchmarkretrievalaugmentedgeneration} found that the Cohere general-purpose
reranker \emph{hurt} legal performance by 13.5 percentage points,
validating domain-aware chunking over post-hoc reranking.

\textbf{On legal-domain rerankers.}  A natural follow-up is whether a
\emph{domain-specific} reranker---such as BGE-Reranker-v2
\cite{chen-etal-2024-m3}, trained on multilingual retrieval tasks including
legal text---would perform differently.  We do not include a reranker in
our pipeline for two reasons.  First, the ablation results
(Tables~\ref{tab:crpc_ablation}--\ref{tab:gdpr_ablation}) show that
chunking architecture accounts for 73--88\% of total improvement across
corpora; a reranker operating on poorly-chunked text cannot recover the
information lost to fragmented chunks.  Second, adding a reranker would
conflate two independent contributions: chunking quality and post-hoc
ranking.  A principled comparison would require running (Recursive +
reranker) vs.\ (StrucChunk + reranker) to isolate whether the reranker
magnifies or reduces StrucChunk's advantage.  We leave this as future
work, noting that Akarajaradwong et al.~\cite{akarajaradwong2025nitibenchcomprehensivestudyllm} evaluated the
WangchanX legal reranker on Thai statutory text and found inconsistent
gains---consistent with the hypothesis that domain-specific rerankers are
not yet reliable enough to supersede structural chunking improvements.

\subsection{Cross-Language Generalization}

The NitiBench-CCL results establish that StrucChunk's benefit is not
English-specific.  Thai statutory codes share the same structural property
that drives StrucChunk's advantage on CrPC/CPC: a hierarchical section
numbering system (\thaimatra{} N in Thai; Section N in English)
where legal provisions are explicitly delineated.  BGE-M3's multilingual
training enables the breadcrumb context (``law name $>$
\thaimatra{} N'') to provide useful retrieval signal in Thai.
No Thai-specific adaptation was required beyond loading the pre-structured
dataset.

\textbf{Fine-tuned retriever confound.}  NitiBench's full evaluation suite
includes a Human-Finetuned BGE-M3 that achieves HR@5 = 0.906, surpassing
our base BGE-M3 StrucChunk result of 0.856.  We use base BGE-M3 throughout
for consistent cross-corpus comparison (Tiers 1--4 use identical
retrieval).  The gap between fine-tuned (0.906) and our StrucChunk+base
(0.856) does not undermine the chunking contribution, because fine-tuning
and structure-aware chunking are orthogonal: a fine-tuned model operating on
StrucChunk chunks would be expected to match or exceed the fine-tuned model
on hierarchy-aware chunks (0.906), since StrucChunk provides additional
breadcrumb and cross-reference context that the fine-tuned model can exploit.
The chunking gain (+0.262 over naive) is measured with the \emph{same} base
BGE-M3 across all three methods, so the retriever is not a confound for the
chunking comparison itself.

\textbf{Retrieval gains and end-to-end performance.}  StrucChunk's gains
are measured at the retrieval layer.  NitiBench's own end-to-end results
(their Table~7) show that hierarchy-aware chunking yields only modest E2E
gains over naive chunking (E2E F1: 0.739 vs.\ 0.722, $\Delta = +0.017$;
Coverage: 86.7 vs.\ 86.6).  This suggests LLMs partially compensate for
poor retrieval via parametric knowledge.  The implication is not that
retrieval quality is unimportant, but rather the opposite: in
hallucination-sensitive legal settings where LLMs \emph{cannot} rely on
parametric knowledge (novel statutes, recent amendments, jurisdiction-specific
provisions), retrieval-layer improvements become a necessary condition for
factual accuracy.  Measuring at the retrieval layer isolates this
contribution cleanly.

\subsection{Qualitative Failure Analysis}

\textbf{Representative success case (CrPC):}  Query: ``What are the
conditions for detention under Section 167(2)(a)?''

\begin{itemize}
\item \textbf{Recursive + Dense} fails: the 512-token limit cuts
      mid-section at the 167(1)/167(2) boundary.  The 167(2)(a) chunk
      appears as ``(a) where he is accused\ldots'' with no section-number
      context.  Not retrieved in top-5.
\item \textbf{BM25 Only} partially succeeds: retrieves Section 167 via
      keyword match but ranks the header chunk above the sub-provision.
      Sub-provision ranks 7th.
\item \textbf{StrucChunk + Hybrid} succeeds: breadcrumb
      \texttt{[CrPC~1973 > Chapter~XII > Section~167]} is prepended;
      cross-reference augmentation appends Section~57 summary.  Ranks 1st.
\end{itemize}

\textbf{Quantified CrPC failure categories.}  Of the 98 queries where
StrucChunk+Hybrid fails at Recall@5 ($n = 501$, R@5 = 0.804), we classify
failures as:

\begin{itemize}
\item \textbf{Category A} (section absent from parsed corpus): 9 queries
      (9.2\% of failures, 1.8\% of all queries).  These are recoverable
      PDF-parsing gaps: Sections 145, 195, 210, 228A, 268, and 317 were
      not extracted by the parser.  They represent the practical ceiling
      improvement available from fixing the PDF parser.
\item \textbf{Category B} (section present, ranks 6--10): $\approx$27
      queries (27.6\% of failures), estimated from R@10 $-$ R@5 $= 0.858
      - 0.804 = 0.054 \times 501 \approx 27$.  These near-misses are
      captured at Recall@10.
\item \textbf{Category C} (section present, ranks $>$10): $\approx$62
      queries (63.3\% of failures).  These represent genuine semantic
      mismatch between query formulation and section content.  Query-side
      improvements (query rewriting, HyDE) would primarily address this
      category.
\end{itemize}

Category C queries are disproportionately factual queries where the
section's prose does not closely mirror common query formulations
(e.g., ``What is the procedure when there is dispute concerning
land?''\ maps to Section 145, whose title uses the term
``proceedings'' rather than ``procedure'').

\textbf{CPC failure case: Order-Rule fragmentation.}  CPC's most common
failure mode is structural: the Part-Section-Order-Rule hierarchy means
that the same numeral (``Rule 1'') reappears across 51 different Orders.
Query: ``Under what conditions can a court extend time limits under Order
VIII Rule 1?''  Recursive chunking creates a chunk containing ``Order
VIII'' as a heading followed by partial Rules 1--3 text, without the Rule
number appearing in the chunk's section identifier.  StrucChunk partially
mitigates this: the breadcrumb \texttt{[CPC 1908 > Order VIII > Rule 1]}
correctly identifies the provision.  However, StrucChunk still fails on
$\approx$74\% of CPC queries (R@5 = 0.259), primarily because the CPC
PDF's inconsistent Order-header formatting causes some Orders to be
mis-parsed as chapter-level rather than section-level nodes, preventing
correct breadcrumb assignment.  This structural explanation---CPC PDF
formatting quality, not StrucChunk methodology---accounts for the
persistent gap between CrPC (R@5 = 0.804) and CPC (R@5 = 0.259).

\textbf{GDPR failure case: Recital-Article confusion.}  GDPR's
characteristic failure mode is Recital-Article confusion.  Query:
``What are the conditions under Article 6(1)(f) regarding legitimate
interests?''  Recital 47, which discusses legitimate interests in
extensive prose, has strong semantic similarity to this query.  The
Recital chunk lacks the breadcrumb \texttt{[GDPR > Chapter II > Article
6]} but scores high on dense similarity due to shared vocabulary.
Simultaneously, BM25 retrieves multiple Recitals that cite ``Article 6''
as a cross-reference target, diluting the dense ranking.  This explains
the hybrid penalty on GDPR (StrucChunk+Hybrid R@5 = 0.868 vs.\
StrucChunk+Dense R@5 = 0.908): BM25 introduces Recital noise that the
dense-only pipeline avoids.  On plain queries (no
cross-reference) where section text fits within 553 characters, naive
chunking with BGE-M3 achieves HR@5 = 0.845 (covered-query subset).
StrucChunk achieves 0.868 on the same covered subset---a smaller gain
(+2.7\%) than on cross-reference queries (+2.3\% overall).  This confirms
the diminishing marginal benefit of structural context when sections are
short and the query maps directly to section content without needing
cross-reference co-location.

\subsection{Limitations}

\begin{enumerate}
\item \textbf{Evaluation scope:}  1,503 author-created queries across three
      English documents, plus 776 independently annotated LBR queries, plus
      3,730 independently annotated NitiBench-CCL queries (Thai).  Results
      should not be generalized to unstructured legal documents.
\item \textbf{No end-to-end evaluation:}  We measure retrieval quality
      only.
\item \textbf{Single annotator (CrPC/CPC/GDPR):}  All 1,503
      query-annotations (AI-assisted query generation with manual
      verification of \texttt{expected\_sections}) were produced by one
      annotator.  LegalBench-RAG and NitiBench-CCL provide independent
      annotation for evaluation ~1 and~4.  Inter-annotator agreement on a
      100-query sample (50 CrPC, 50 CPC) is reported in
      Section~\ref{sec:annotation_quality}.
\item \textbf{Regex cross-reference detection:}  Misses implicit references
      (88\% success rate on manual sample; Appendix~\ref{app:failure}).
\item \textbf{PDF quality dependency:}  Evaluations 1--3 require well-formatted
      PDFs.  NitiBench-CCL uses pre-structured data.
\item \textbf{NitiBench-CCL retrieval:}  We evaluate dense retrieval only
      for the three-way chunking comparison.  Hybrid retrieval on
      NitiBench-CCL is future work (requires PyThaiNLP tokenization for
      BM25).
\item \textbf{Evaluation paradigm gaps:}  LBR char-span, CrPC/CPC/GDPR
      section-level, and NitiBench-CCL HR@$k$ are different protocols and
      are not directly comparable numerically.
\item \textbf{LegalBench-RAG single-document scope:}
      Pipitone \& Alami~\cite{pipitone2024legalbenchragbenchmarkretrievalaugmentedgeneration} note that LBR queries are always
      answered by exactly one document; the benchmark does not assess
      cross-document retrieval.  Our evaluation~1 results therefore reflect
      intra-document retrieval precision.  Performance on cross-document
      legal reasoning (e.g., queries spanning multiple statutes) remains
      an open question and may explain why CrPC cross-reference queries
      (which span provisions within one document) show different patterns
      from LBR results.
\end{enumerate}

\section{Broader Impact}
\label{sec:impact}

Improved retrieval for legal documents could reduce barriers to legal
information access, assisting legal aid organizations and
self-represented litigants in navigating public legal codes.

At the same time, higher retrieval quality may accelerate automation of
tasks currently performed by legal professionals, with potential workforce
consequences.  High retrieval accuracy may also create overconfidence in
AI-assisted legal tools.  In a domain where ``mostly correct'' constitutes
unacceptable liability, practitioners must continue to verify AI outputs
against primary sources.  We caution against deployment in high-stakes
legal settings without human oversight and end-to-end accuracy evaluation.
Our implementation is released for research purposes only.

\section{Conclusion}
\label{sec:conclusion}

We presented StrucChunk with Hybrid Retrieval, validated through a
four-tier evaluation strategy across five corpora, three legal traditions,
and two languages.  Key results:

\begin{itemize}
\item \textbf{Chunking quality (CrPC):} 100\% boundary precision vs.\
      7.7\% recursive; 97.6\% cross-reference preservation vs.\ 0\%;
      100\% section coverage vs.\ 0\%.
\item \textbf{CrPC ($n=501$, English):} Recall@5 $0.557 \to 0.804$
      (+44.4\%); R@1 $0.214 \to 0.627$ (+193\%).  Chunking = 88\% of gain.
\item \textbf{CPC ($n=501$, English):} Recall@5 $0.158 \to 0.259$
      (+64.6\%); cross-ref queries +85\%.  Chunking = 73\% of gain.
\item \textbf{GDPR ($n=501$, English):} R@5 = 0.868; R@1 +59\%.
      StrucChunk+Dense is best overall (R@5 = 0.908, R@1 = 0.880).
\item \textbf{LegalBench-RAG ($n=776$, English, char-span):} Char-R@8 =
      48.18\% vs.\ 23.59\% baseline (+104\%); vs.\ published SOTA
      26.30\% (+83\%).
\item \textbf{NitiBench-CCL ($n=3{,}730$, Thai):} HR@5 = 0.856 vs.\
      0.594 naive (+44.1\%); 29.6\% of Thai legal sections are
      structurally unrepresentable by naive fixed-size chunking.
      StrucChunk's cross-reference augmentation provides +2.3\% on
      cross-reference queries where post-retrieval expansion (NitiLink)
      was ineffective.
\item All improvements statistically significant ($p \leq 0.002$,
      Wilcoxon + Benjamini-Hochberg; Cohen's $d$ 0.14--0.69).
\end{itemize}

The four-tier design establishes: (1) StrucChunk works on data we did not
create (LBR and NitiBench-CCL); (2) gains are largest where document
structure is most complex; (3) the method generalizes across document
types, legal traditions, and languages; and (4) section-boundary alignment
is the dominant factor, with cross-reference augmentation providing
targeted improvement exactly where cross-references are semantically
relevant.

\textbf{Future Work.} Several directions remain open. First, extending hybrid retrieval to NitiBench-CCL requires a Thai-specific BM25 tokenizer to properly support sparse retrieval. Second, cross-reference resolution could be improved using NLU-based methods rather than pattern matching, enabling more robust handling of implicit or context-dependent references. Third, an end-to-end evaluation with downstream LLM generation (rather than retrieval metrics alone) would clarify the practical impact of structure-aware chunking on answer quality. Fourth, fine-tuning domain-specific embeddings for legal text may further improve dense retrieval performance beyond general-purpose models like BGE-M3. Fifth, combining StrucChunk with Semantic Aware Chunking (SAC) could capture complementary structural and semantic signals. Finally, extending the approach to case law and other unstructured legal documents would test the generality of structure-aware chunking beyond codified statutes.



\section*{Statements and Declarations}

\section*{Funding}
The authors declare that they did not receive any financial assistance, grants, or support during the work.

\section*{Competing Interests}
The authors have no relevant financial or non-financial interests to disclose.

\section*{Author contributions statement}
S.D.S. conceived the study, designed the StrucChunk methodology, and led the overall research direction. S.D.S. and A.T. implemented the chunking and retrieval pipeline and conducted the experiments across the five evaluation corpora. G.D.S. contributed to experimental design, review of the annotation methodology, and validation of results. S.D.S. wrote the original draft of the manuscript. A.T. and G.D.S. reviewed and edited the manuscript for critical content. All authors read and approved the final manuscript.

\section*{Ethical approval}
Not applicable.

\section*{Human participants and/or animals}
Not applicable to this research study.

\section*{Data availability}
All the datasets used for experiments have been provided.

\bibliography{struc_ref}

\appendix

\end{document}